\documentclass[letterpaper]{article}
\usepackage{proceed2e}
\usepackage[margin=1in]{geometry}
\usepackage{hyperref}
\usepackage{url}

\usepackage{amsmath,blkarray,bbold,graphicx}
\usepackage{float}
\usepackage{times}
\usepackage{graphicx} 
\usepackage{subfigure} 

\usepackage{algorithm}
\usepackage{algorithmic}
\usepackage{amsmath,blkarray,bbold,graphicx}
\usepackage[colorinlistoftodos]{todonotes}
\usepackage{url}
\usepackage{amsmath}
\usepackage{times}
\usepackage{import}
\usepackage{natbib} %
\usepackage{hyperref} 
\usepackage{url} %
\usepackage{fancyvrb}
\usepackage{subfigure} %
\usepackage{comment} %
\usepackage{mdframed}
\usepackage{enumitem}
\usepackage{relsize}
\usepackage{framed}
\usepackage{graphicx} %
\usepackage{wrapfig}
\usepackage[colorinlistoftodos]{todonotes}
\usepackage{multirow}
\usepackage[capitalize]{cleveref}

\usepackage{array}
\newcolumntype{L}[1]{>{\raggedright\let\newline\\\arraybackslash\hspace{-8pt}}m{#1}}
\newcolumntype{C}[1]{>{\centering\let\newline\\\arraybackslash\hspace{-12pt}}m{#1}}
\newcolumntype{R}[1]{>{\raggedleft\let\newline\\\arraybackslash\hspace{-8pt}}m{#1}}
\usepackage{changepage}

\crefname{nlem}{Lemma}{Lemmas}
\crefname{nprop}{Proposition}{Propositions}
\crefname{ncor}{Corollary}{Corollaries}
\crefname{nthm}{Theorem}{Theorems}
\crefname{assumption}{Assumption}{Assumptions}

\usepackage{misc}
\title{Multimodal Prediction and Personalization of Photo Edits \\ with Deep Generative Models}

\author{ Ardavan Saeedi \\
CSAIL, MIT
\And 
Matthew D. Hoffman \\
Adobe Research 
\And 
Stephen J. DiVerdi \\
Adobe Research 
\AND Asma Ghandeharioun \\
Media Lab, MIT 
\And Matthew J. Johnson \\
Google Brain
\And Ryan P. Adams\\
Harvard and Google Brain}

\begin{document}

\maketitle
%

%



\begin{abstract}
Professional-grade software applications are powerful but complicated---expert users can achieve impressive results, but novices often struggle to complete even basic tasks.
Photo editing is a prime example: after loading a photo, the user is confronted with an array of cryptic sliders like ``clarity'', ``temp'', and ``highlights.''
An automatically generated suggestion could help, but there is no single ``correct'' edit for a given image---different experts may make very different aesthetic decisions when faced with the same image, and a single expert may make different choices depending on the intended use of the image (or on a whim).
We therefore want a system that can propose multiple diverse, high-quality edits while also learning from and adapting to a user's aesthetic preferences.
In this work, we develop a statistical model that meets these objectives.
Our model builds on recent advances in neural network generative modeling and scalable inference, and uses hierarchical structure to learn editing patterns across many diverse users.
Empirically, we find that our model outperforms other approaches on this challenging multimodal prediction task.
\end{abstract}

\section{Introduction}
Many office workers spend most of their working days using pro-oriented software applications. These applications are often powerful, but complicated. This complexity may overwhelm and confuse novice users, and even expert users may find some tasks time-consuming and repetitive. We want to use machine learning and statistical modeling to help users manage this complexity.

Fortunately, modern software applications collect large amounts of data from users with the aim of providing them with better guidance and more personalized experiences. A photo-editing application, for example, could use data about how users edit images to learn what kinds of adjustments are appropriate for what images, and could learn to tailor its suggestions to the aesthetic preferences of individual users. Such suggestions can help both experts and novices: experts can use them as a starting point, speeding up tedious parts of the editing process, and novices can quickly get results they could not have otherwise achieved.

Several models have been proposed for predicting and personalizing user interaction in different software applications.

These existing models are limited in that they only propose a single prediction or are not readily personalized. Multimodal predictions\footnote{We mean ``multimodal'' in the statistical sense (i.e., coming from a distribution with multiple maxima), rather than in the human-computer-interaction sense (i.e., having multiple modes of input or output).} are important in cases where, given an input from the user, there could be multiple possible suggestions from the application. For instance, in photo editing/enhancement, a user might want to apply different kinds of edits to the same photo depending on the effect he or she wants to achieve. A model should therefore be able to recommend multiple enhancements that cover a diverse range of styles.

\begin{figure*}[t!]
\begin{center}
\begin{tabular}{cc}
{\includegraphics[trim = 0mm -10mm 0mm 0mm, clip, scale = 0.23]{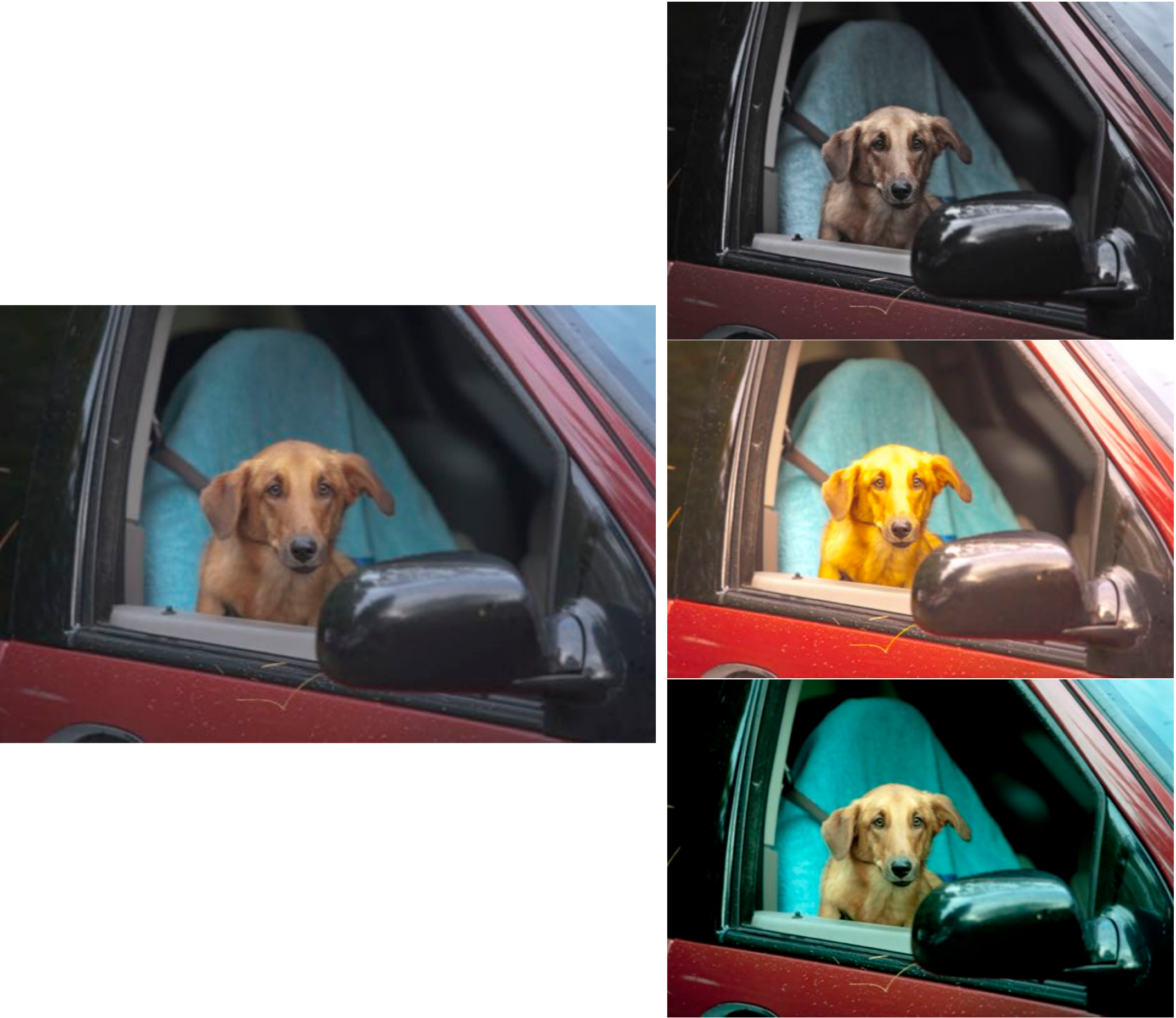}} & \hspace{5mm}
{\includegraphics[trim = 0mm 0mm 0mm 0mm, clip, scale = 0.25]{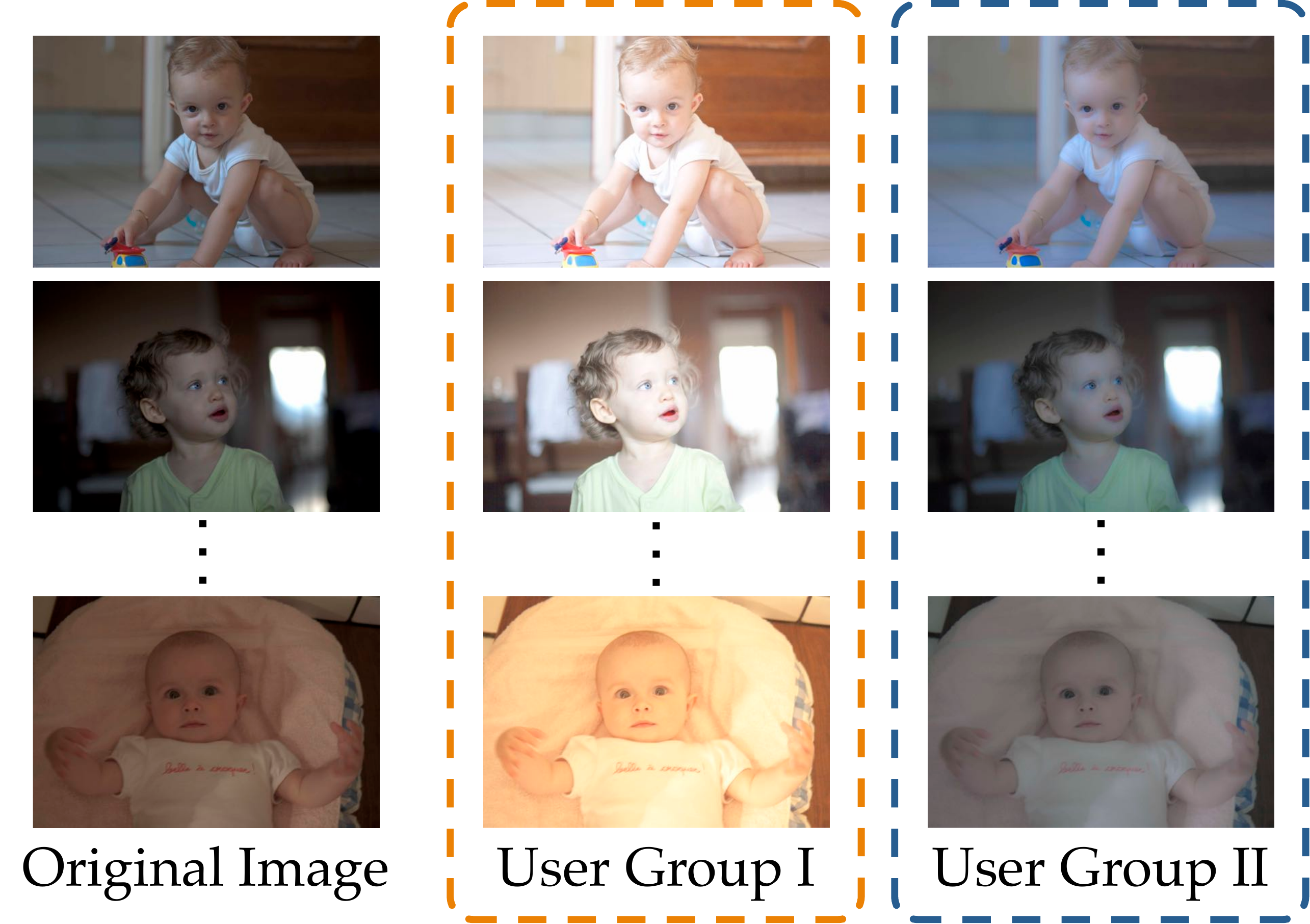}} \\
(a) & (b)
\end{tabular}
\end{center}
\vspace{-4mm}
\caption{\footnotesize{The main goals of our proposed models: (a) \textbf{Multimodal photo edits}: For a given photo, there may be multiple valid aesthetic choices that are quite different from one another. (b) \textbf{User categorization}: A synthetic example where different user clusters tend to prefer different slider values. Group 1 users prefer to increase the exposure and temperature for the baby images; group 2 users reduce clarity and saturation for similar images.}}
\label{fig:synthetic_example}
\end{figure*}

In this paper, we introduce a framework for multimodal prediction and personalization in software applications. We focus on photo-enhancement applications, though our framework is also applicable to other domains where multimodal prediction and personalization is valuable. Figure \ref{fig:synthetic_example} demonstrates our high-level goals: we want to learn to propose diverse, high-quality edits, and we want to be able to personalize those proposals based on users' historical behavior.

Our modeling and inference approach is based on the variational autoencoder (VAE) \cite{kingma2013auto} and a recent extension of it, the structured variational autoencoder (SVAE) \cite{johnson2016structured}. Along with our new models, we develop approximate inference architectures that are adapted to our model structures.

We apply our framework to three different datasets (collected from novice, semi-expert, and expert users) of image features and user edits from a photo-enhancement application and compare its performance qualitatively and quantitatively to various baselines. We demonstrate that 
our model outperforms other approaches.

\section{Background and related work}
In this section, we first briefly review the frameworks (VAEs and SVAEs) that our model is built upon; next, we provide an overview of the available models for predicting photo edits and summarize their pros and cons. 

\subsection{Variational autoencoder (VAE)}
\label{sec:vae}
The VAE, introduced in \cite{kingma2013auto}, has been successfully applied to various models with continuous latent variables and a complicated likelihood function (e.g., a neural network with nonlinear hidden layers). In these settings, posterior inference is typically intractable, and even approximate inference may be prohibitively expensive to run in the inner loop of a learning algorithm. The VAE allows this difficult inference to be amortized over many learning updates, making each learning update cheap even with complex likelihood models.

As an instance of such models, consider modeling a set of~$N$ i.i.d.\ observations~${y = \{y_n\}^N_{n=1}}$ with the following generative process:~${z_n \overset{\text{iid}}\sim h}$ and~${y_n \sim f(g_\theta(z_n))}$, where~$z_n$ is a latent variable generated from a prior~$h(z)$ (e.g.,~$\mathcal{N}(0, \it{I})$) and the likelihood function~${p_{\theta}(y_n| z_n) = f(y_n; g_\theta(z_n))}$ is a simple distribution $f$ whose parameters $g_\theta(z_n)$ can be a complicated function of $z_n$. For example, $p_\theta(y_n|z_n)$ might be~$\mathcal{N}(y_n; \mu(z_n;\theta), \Sigma(z_n;\theta))$ where the mean and the covariance depend on~$z_n$ through a multi-layer perceptron (MLP) richly parameterized by weights and biases~$\theta$. See Figure~\ref{fig:vae_svae}(a) for the graphical model representation of this generative process.

In the VAE framework, the posterior density~$p_{\theta}(z|y)$ is approximated by a recognition network~$q_{\phi}(z|y)$, which can take the form of a flexible conditional density model such as an MLP parameterized by~$\phi$. To learn the parameters of the likelihood function~$\theta$ and the recognition network~$\phi$, the following lower bound on the marginal likelihood is maximized with respect to~$\theta$ and~$\phi$:
\[
\label{eq:vae}
\mathcal{L}_{\text{VAE}}(\phi, \theta) \triangleq \mathbb{E}_{q_{\phi}(z|y)}[\text{log} \ p_{\theta}(y|z)]-\text{KL}(q_{\phi}(z|y)||p(z)) .
\]
To compute a Monte Carlo estimate of the gradient of this objective with respect to~$\phi$, \citet{kingma2013auto} propose a reparameterization trick for sampling from~$q_{\phi}(z|y)$ by first sampling from an auxiliary noise variable and then applying a differentiable map to the sampled noise. This yields a differentiable Monte Carlo estimate of the expectation with respect to~$\phi$. Given the gradients, the parameters are updated by stochastic gradient ascent.    

\subsection{Structured variational autoencoder (SVAE)}

\label{sec:svae}
\begin{figure}[t!]
\begin{center}
\begin{tabular}{cc}
{\includegraphics[trim = 0mm 0mm 0mm 0mm, clip, scale = 0.35]{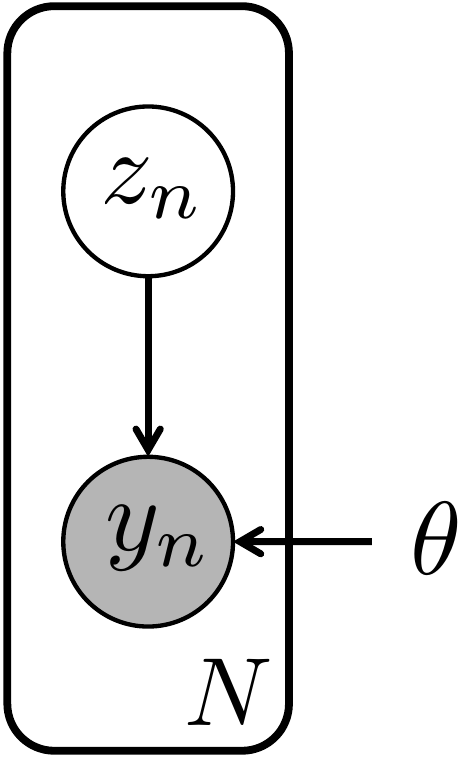}} & \ \ \qquad
\includegraphics[trim = 0mm 0mm 0mm 0mm, clip, scale = 0.35]{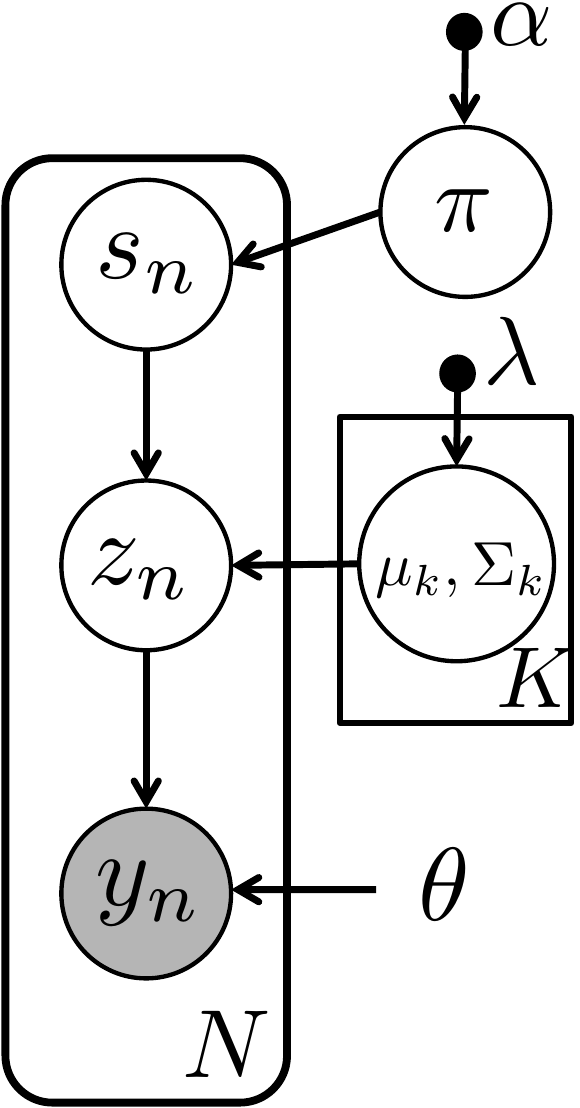}\\
(a) & (b)
\end{tabular}
\end{center}
\vspace{-4mm}
\caption{\footnotesize{(a) VAE with Gaussian latent variables (Section~\ref{sec:vae}) (b) SVAE with a GMM latent graphical model (Section~\ref{sec:svae})}}

\label{fig:vae_svae}
\end{figure}

\cite{johnson2016structured} extend the VAE inference scheme to latent graphical models with neural network observation distributions. This SVAE framework combines the interpretability of graphical models with the flexible representations found by deep learning. For example, consider a latent Gaussian mixture model (GMM) with nonlinear observations:
\[
\nonumber
\pi \sim \text{Dir}(\alpha), \quad (\mu_k, \Sigma_k) &\sim \text{NIW}(\lambda), \quad s_n \overset{\text{iid}}\sim \pi, \\
 \quad z_n|s_n, \{\mu_k, \Sigma_k\}^{K}_{k=1} &\sim \mathcal{N}(\mu_{s_n}, \Sigma_{s_n}), \\ 
 \nonumber
 y_n | z_n, \theta &\sim \mathcal{N}(\mu(z_n;\theta), \Sigma(z_n;\theta))\,.
\]
Note that the nonlinear observation model for each data point $y_n | z_n, \theta$ resembles that of the VAE, while the latent variable model for $z_n$ is a GMM (see Figure \ref{fig:vae_svae}(b)).
This latent GMM can represent explicit latent cluster assignments while also capturing complex non-Gaussian cluster shapes in the observations.

To simplify the SVAE notation, we consider a general setting in which we denote the global parameters of a graphical model by~$\gamma$ and the local latent variables by~$z$.
Furthermore, we assume that~$p(z|\gamma)$ and~$p(\gamma)$ are a conjugate pair of exponential family densities with sufficient statistics~$t_{z}(z)$ and~$t_{\gamma}(\gamma)$.
We continue to use $\theta$ to denote the parameters of a potentially complex, nonlinear observation likelihood.
Using a mean field family distribution for approximating the posterior the variational lower bound (VLB) can be written as:
\[
\label{eq:SVAE1}
\mathcal{L}(\eta_{\gamma}, \theta, \eta_{z}) \triangleq \mathbb{E}_{q(\gamma)q(z)}\left[\log \frac{p(\gamma)p(z|\gamma)p_{\theta}(y|z)}{q(\gamma)q(z)} \right] 
\]
where $\eta_{\gamma}$ and $\eta_{z}$ are the parameters of the variational distributions~$q(\gamma)$ and~$q(z)$ respectively. 

Due to the non-conjugate likelihood function $p_\theta(y | z)$, standard variational inference methods cannot be applied to the latent GMM. To solve this problem, the SVAE replaces the non-conjugate likelihood with a recognition model $r(y;\phi)$ that generates conjugate evidence potentials. We can then define a surrogate objective~$\widehat{\mathcal{L}}$ with conjugacy structure:
\[
\label{eq:svaesurr}
\widehat{\mathcal{L}}(\eta_{\gamma}, \eta_{z}, \phi) &\triangleq \\
\nonumber
\mathbb{E}_{q(\gamma)q(z)}&\left[\log \frac{p(\gamma)p(z|\gamma)\exp\{\langle r(y;\phi), t_z(z)\rangle\}}{q(\gamma)q(z)} \right]  
\]
where the potentials~$\exp \{ \langle r(y;\phi), \, t_z(z)\rangle \}$ have a conjugate form to~$p(z|\gamma)$.
By choosing $\eta_z$ to optimize this surrogate objective, writing ${\eta^*_{z}(\eta_{\gamma}, \phi) \triangleq \operatorname{arg\,max}_{\eta_z} \widehat{\mathcal{L}}(\eta_{\gamma}, \eta_{z}, \phi)}$, the SVAE objective is then~${\mathcal{L}_{\text{SVAE}}(\eta_{\gamma}, \theta, \phi)\triangleq \mathcal{L}(\eta_{\gamma}, \theta, \eta^*_{z}(\eta_{\gamma}, \phi))}$ which can be shown to lower bound the variational inference objective in eq.~\ref{eq:SVAE1}. As in the stochastic variational inference (SVI) algorithm \cite{hoffman2013stochastic}, there is a simple expression for the natural gradient of this objective with respect to  the variational parameters with conjugate priors;
the gradients w.r.t. other variational parameters, such as those parameterizing neural networks, can be computed using the reparameterization trick.

\subsection{Related work on the prediction of photo edits}
\label{sec:relatedwork}
There are two main categories of models, parametric and nonparametric, that have been used for prediction of photo edits:

\paragraph{Parametric methods} These methods approximate a parametric function by minimizing a squared (or a similar) loss. The loss is typically squared~$L_2$ distance in Lab color space, which more closely approximates human perception than RGB space \citep{sharma2002digital}. This loss is reasonable if the goal is to learn from a set of consistent, relatively conservative edits. But when applied to a dataset of more diverse edits, a model that minimizes squared error will tend to predict the \emph{average} edit. At best, this will lead to conservative predictions; in the worst case, the average of several good edits may produce a bad result.

\citet{bychkovsky2011learning} collect a dataset of 5000 photos enhanced by 5 different experts; they identify a set of features and learn to predict the user adjustments after training on the collected dataset. They apply a number of regression techniques such as LASSO and Gaussian-process regression and show their proposed adjustments can match the adjustments of one of the 5 experts. Their method only proposes a single adjustment and the personalization scheme that they suggest requires the user to edit a set of selected training photos.  

\citet{yan2016automatic} use a deep neural network to learn a mapping from an input photo to an enhanced one following a particular style; their results show that the proposed model is able to capture the nonlinear and complex nature of this mapping. They also incorporate semantic awareness in their model, so their model can predict the adjustments based on the semantically meaningful objects (e.g., human, animal, sky, etc.) in the photo. This method also only proposes a single style of adjustment.

\citet{jaroensri2015predicting} propose a technique that can predict an acceptable range  of adjustments for a given photo. The authors crowd-sourced a dataset of photos with various brightness and contrast adjustments, and asked the participants to label each edited image as ``acceptable'' or ``unacceptable". 

From this labeled dataset they learn a support vector machine classifier that can determine whether an adjustment is acceptable or not. They use this model to predict the acceptable range of edits by first sampling from the parameter space and then using their learned model to analyze each sample. Although their model is able to propose a range of edits to the user, it requires a balanced, human-labeled training set of ``acceptable" and ``unacceptable" images. Since the number of bad edits may grow exponentially with the dimensionality of the adjustment space, they mostly limit their study to two-dimensional brightness and contrast adjustments.

\paragraph{Nonparametric methods} These methods are typically able to propose multiple edits or some uncertainty over the range of adjustments. 

\citet{lee2015automatic} propose a method that can generate a diverse set of edits for an input photograph. The authors have a curated set of exemplar images in various styles. They use an example-based style-transfer algorithm to transfer the style from an exemplar image to an input photograph. To choose the right exemplar image, they do a semantic similarity search using features that they have learned via a convolutional neural network (CNN). Although their approach can recommend multiple edits to a photo, their edits are destructive; that is, the user is not able to customize the model's edits.

\citet{koyama2016selph} introduce a model for personalizing photo edits only based on the history of edits by a single user. The authors use a self-reinforcement procedure in which after every edit by a user they~1)~update the distance metric between the user's past photos~2)~update a feature vector representation of the user's photos and~3)~update an enhancement preference model based on the feature vectors and the user's enhancement history. This model requires data collection from a single user and does not benefit from other users' information.

\subsection{Related multimodal prediction models}
Traditional neural networks using mean squared error (MSE) loss cannot naturally handle multimodal prediction problems, since MSE is minimized by predicting the average response. \citet{neal1992connectionist} introduces stochastic latent variables to the network and proposes training Sigmoid Belief Networks (SBN) with only binary stochastic variables. However, this model is difficult to train, and it can only make piecewise-constant predictions and is therefore not a natural fit to continuous-response prediction problems.

\citet{bishop1994mixture} proposes mixture density networks (MDN), which are more suitable for continuous data. Instead of using stochastic units, the model directly outputs the parameters of a Gaussian mixture model. That is, a some of the network outputs are used as mixing weights and the rest provide the means and variances of the mixture components. The complexity of MDNs' predictive distributions is limited by the number of mixture components if the optimal predictive distribution cannot be well approximated by a relatively small number of Gaussians, then an MDN may not be an ideal choice.

\citet{tang2013learning} add deterministic hidden variables to SBNs in order to model continuous distributions. The authors showed improvements over the SBN; nevertheless, training the stochastic units remained a challenge due to the difficulty of doing approximate inference on a large number of discrete variables.

\citet{dauphin2015predicting} propose a new class of stochastic networks called linearizing belief networks (LBN). LBN combines deterministic units with stochastic binary units multiplicatively. The model uses deterministic linear units which act as multiplicative skip connections and allow the gradient to flow without diffusion. The empirical results show that this model can outperform standard SBNs.

\section{Models}

Given the limitations of the available methods for predicting photo edits (described in Section~\ref{sec:relatedwork}), our goal is to propose a framework in which we can: 1) recommend a set of diverse, parametric edits based on a labeled dataset of photos and their enhancements, 2) categorize the users based on their style and type of edits they apply, and finally 3) personalize the enhancements based on the user category. We focus on the photo-editing application in this paper, but the proposed framework is applicable to other domains where users must make a selection from a large, richly parameterized design space where there is no single right answer (for example, many audio processing algorithms have large numbers of user-tunable parameters).

Our framework is based on VAEs and their extension SVAEs, and follows a mixture-of-experts design \citep[Section 11.2.4]{murphy2012machine}. We first introduce a conditional VAE that can generate diverse set of enhancements to a  given photo. Next, we extend the model to categorize the users based on their adjustment style. Our model can provide interpretable clusters of users with similar style. Furthermore, the model can provide personalized suggestions by first estimating a user's category and then suggesting likely enhancements conditioned on that category.  

\subsection{Multimodal prediction with conditional Gaussian mixture variational autoencoder (CGM-VAE)}
\label{sec:cvae}
Given a photo, we are interested in predicting a set of edits. Each photo is represented by a feature vector $x_n$ and its corresponding edits $y_n$ are represented by a vector of slider values (e.g. contrast, exposure, saturation, etc.). We assume that there are $L$ clusters of possible edits for each image. To generate the sliders $y_n$ for a given image~$x_n$, we first sample a cluster assignment $s_n$ and a set of latent features $z_n$ from its corresponding mixture component $\mathcal{N}(\mu_{s_n}, \Sigma_{s_n})$. Next, conditioned on the image and~$z_n$, we sample the slider values. The overall generative process for the slider values $\{y_n\}_{n=1}^N$ conditioned on the input images $\{x_n\}_{n=1}^N$ is
\[
\nonumber
s_n|\pi \overset{\text{iid}}\sim \pi, \ \ \ \ \ \ \
z_n|s_n, &\{\mu_{\ell}, \Sigma_{\ell}\}^L_{\ell=1} \sim \mathcal{N}(\mu_{s_n}, \Sigma_{s_n}), \\
y_n | x_n, z_n, \theta \sim \mathcal{N}(&\mu(z_n, x_n;\theta), \Sigma(z_n, x_n;\theta)),
\label{eq:cvae}
\]
where $\mu(z_n, x_n;\theta)$ and $\Sigma(z_n, x_n;\theta)$ are flexible parametric functions, such as MLPs, of the input image features $x_n$ concatenated with the latent features $z_n$. Summing over all possible values for the latent variables $s_n$ and $z_n$, the marginal likelihood $p(y_n|x_n)=\sum_{s_n}\int_{z_n} p(y_n, s_n, z_n|x_n)dz_n$ yields complex, multimodal densities for the image edits $y_n$.

The posterior $p(s, z | x, y)$ is intractable. We approximate it with variational recognition models as
\begin{equation}
  p_{\theta}(s, z|x, y) \approx q_{\phi_s}(s|x, y) q_{\phi_z}(z|x, y, s).
\end{equation}
Note that this variational distribution does not model $s$ and $z$ as independent. For $q_{\phi_s}(s|x, y)$, we use an MLP with a final softmax layer, and for $q_{\phi_z}(z|x, y, s)$, we use a Gaussian whose mean and covariance are the output of an MLP that takes $s$, $x$, and $y$ as input.

Given this generative model and variational family, to perform inference we maximize a variational lower bound on $\log p_{\theta}(y|x)$, writing the objective as
\[
\nonumber
\mathcal{L}(\theta, \phi) \triangleq
&\mathbb{E}_{q_{\phi}(s, z|x,y)} \! [\log p_{\theta}(y|z,x)] \\ 
&- \text{KL}(q_{\phi}(s, z|x,y)||p_{\theta}(s, z)).
\nonumber
\]
By marginalizing over the latent cluster assignments $s$, the CGM-VAE objective can be optimized using stochastic gradient methods and the reparameterization trick as in Section~\ref{sec:vae}. Marginalizing out the discrete latent variables is not computationally intensive since $s$ and $y$ are conditionally independent given $z$, $p_\theta(s, z)$ is cheap to compute relative to $p_\theta(y|x, z)$, and we use a relatively small number of clusters. However, with a very large discrete latent space, one could use alternate approaches such as the Gumbel-Max trick \cite{maddison2016concrete}.

\begin{figure*}[t!]
\begin{center}
\begin{tabular}{cccc}
{\includegraphics[trim = 0mm 0mm 0mm 0mm, clip, scale = 0.35]{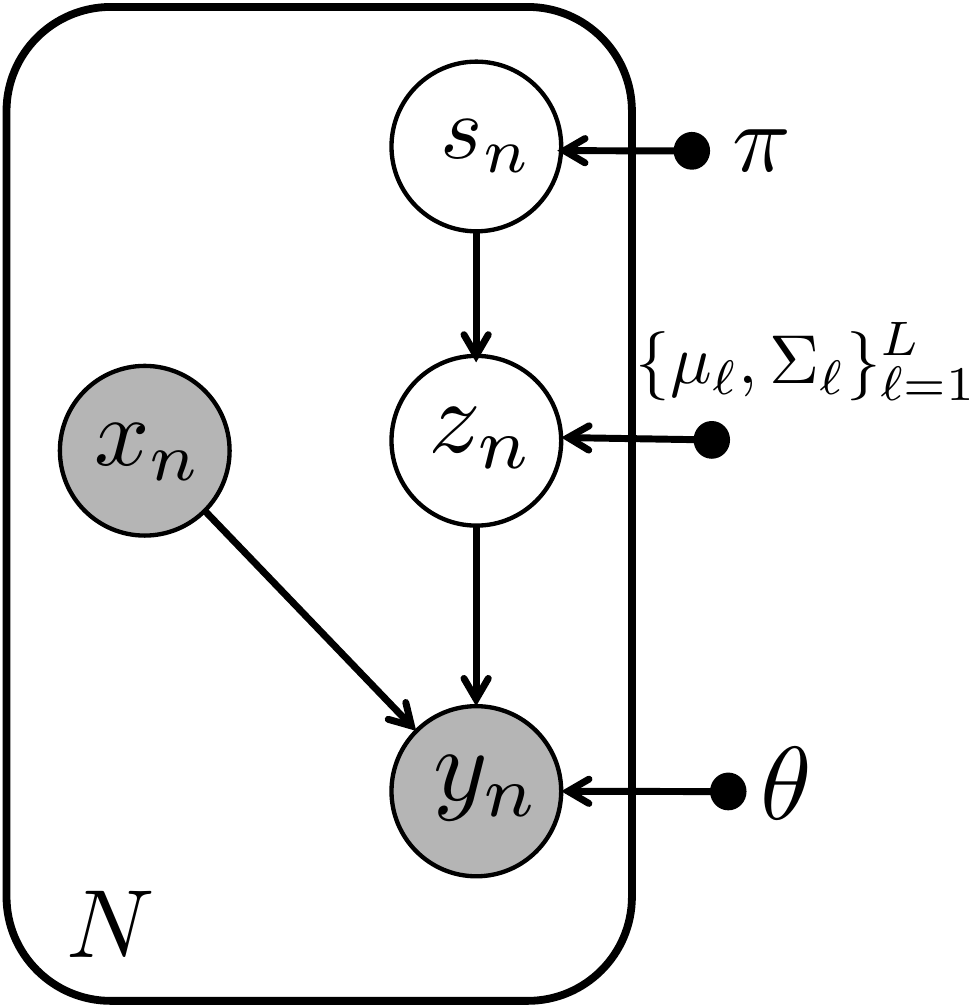}} & \ \ 
\includegraphics[trim = 0mm 0mm 0mm 0mm, clip, scale = 0.35]{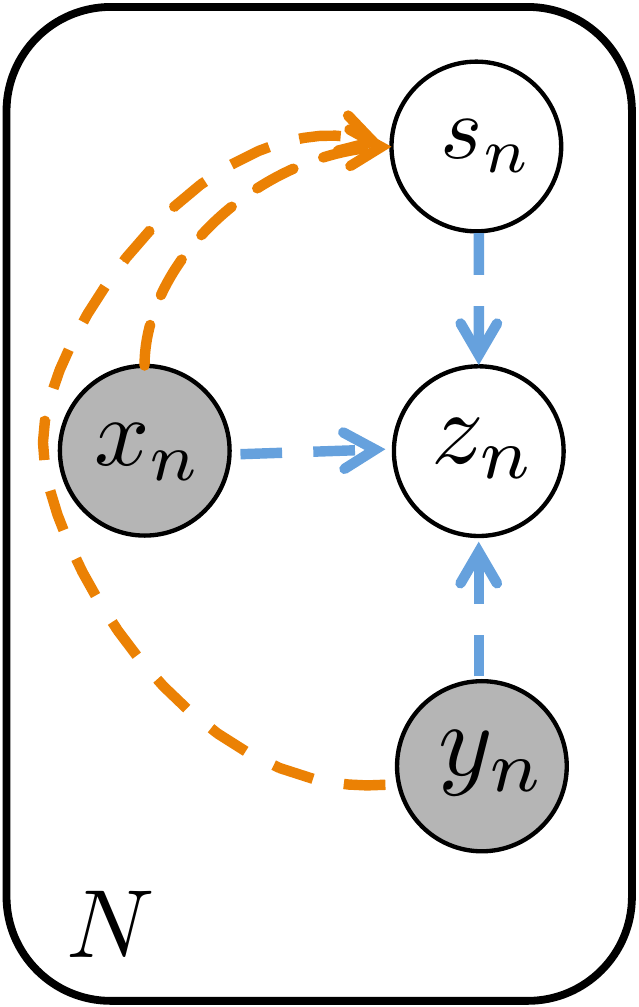}  & \ \ \qquad
\includegraphics[trim = 0mm 0mm 0mm 0mm, clip, scale = 0.35]{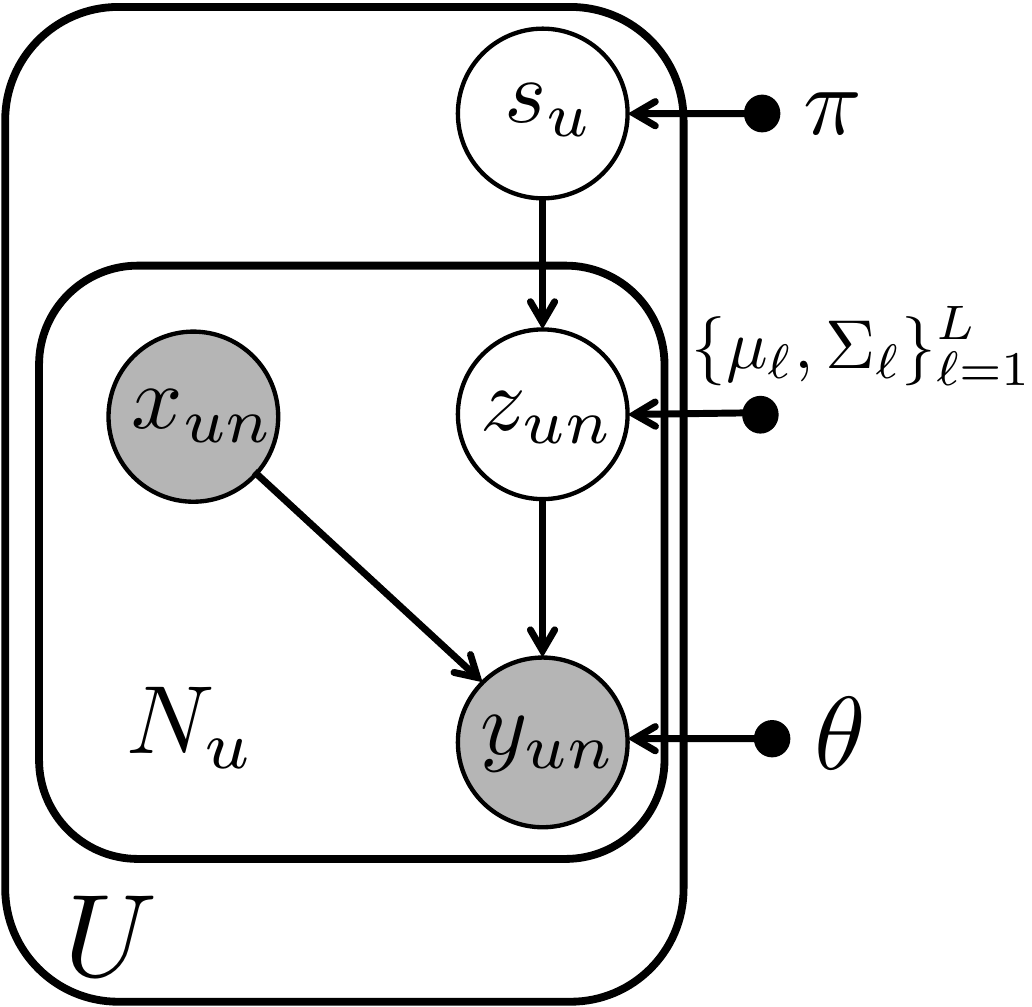}  \ & \ \ 
\includegraphics[trim = 0mm 0mm 0mm 0mm, clip, scale = 0.35]{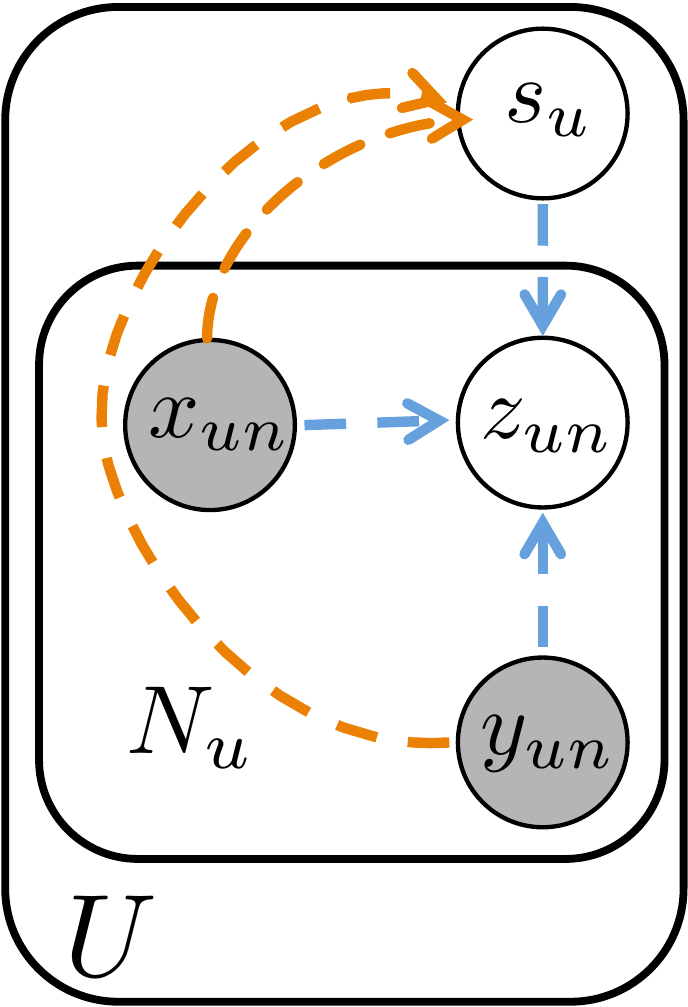}\\
\hspace{-12mm}(a) & \hspace{2mm}(b) & \hspace{-2mm}(c) & (d)
\end{tabular}
\end{center}
\vspace{-4mm}
\caption{\footnotesize{(a) The graphical model for CGM-VAE introduced in Section \ref{sec:cvae} (b) The dependency structure for the variational approximations $q_{\phi_s}(s|x,y)$ and $q_{\phi_z}(z|x,y,s)$ in CGM-VAE (c) The CGM-SVAE model introduced in Section~\ref{sec:personalization}) for categorization and personalization. There are $U$ users and each user $u$ has $N_u$ photos. (d) The dependency structure in the variational distributions for the CGM-SVAE model. Note that the recognition network for $s_u$ depends on all the images and their corresponding slider values of user $u$.}}

\label{fig:cvae_graphical_models}
\end{figure*}

Figure~\ref{fig:cvae_graphical_models} (parts a and b) outlines the graphical model structures of the CGM-VAE and its variational distributions $q_\phi$.

\subsection{Categorization and personalization}
\label{sec:personalization}
In order to categorize the users based on their adjustment style, we extend the basic CGM-VAE model to a hierarchical model that clusters users based on the edits they make. While the model in the previous section considered each image-edit pair $x_n, y_n$ in isolation, we now organize the data according to $U$ distinct users, using $x_{un}$ to denote the $n$th image of user $u$ and $y_{un}$ to denote the corresponding slider values. $N_u$ denotes the number of photos edited by user $u$. 
As before, we assume a GMM with $L$ components $\{\mu_{\ell}, \Sigma_{\ell}\}^L_{\ell=1}$ and mixing weights $\pi$ to model the user categories.

For each user $u$ we sample a cluster index $s_u$ to indicate the user's category, then for each photo $n\in\{1,\ldots,N_u\}$ we sample the latent attribute vector $z_{un}$ from the corresponding mixture component:  
\[
\nonumber
s_u | \pi \overset{\text{iid}}\sim \pi, \quad  z_{un} | s_u, \{(\mu_{\ell}, \Sigma_{\ell})\}_{\ell=1}^L \overset{\text{iid}}\sim \mathcal{N}(\mu_{s_u}, \Sigma_{s_u}).
\]
Finally, we use the latent features $z_{un}$ to generate the vector of suggested slider values $y_{un}$. As before, we use a multivariate normal distribution with mean and variance generated from an MLP parameterized by $\theta$:
\[
\nonumber
y_{un} | x_{un}, z_{un},  \theta \overset{\text{iid}}\sim \mathcal{N}(\mu(z_{un}, &x_{un};\theta), \Sigma(z_{un}, x_{un};\theta)).
\]

For inference in the CGM-SVAE model, our goal is to maximize the following VLB:
\[
\label{eq:SVAE2}
\mathcal{L}\triangleq \mathbb{E}_{q}\left[\log \frac{p(\textbf{y},\textbf{z},\textbf{s} | \textbf{x})}{q(\textbf{s, z})} \right].
\]
To optimize this objective, we follow a similar approach to the SVAE inference framework described in Section~\ref{sec:svae}. In the following we define the variational factors and the recognition networks that we use.

\paragraph{Variational factors} 

For the local variables $z$ and $s$, we restrict $q(z|s)$ to be normal with natural parameters $\eta_z$ and we have $q(s)$ in the categorical form with natural parameter $\eta_s$. As in the CGM-VAE, we marginalize over cluster assignments at the user level. 

For a dataset of $U$ users, the VLB factorizes as follows:
\vspace{-2mm}
\[
\label{eq:csvae}
\nonumber
\mathcal{L}(\theta, \phi) \triangleq \frac{1}{U}&\sum_u \Big[\sum^{N_u}_{n=1} \mathbb{E}_{q_{\phi}(z, s|x,y)}[\log p_{\theta} (y_{un}| x_{un}, {z}_{un})] \\
\nonumber
&\ \ - \text{KL}(q_{\phi}(z_{un}|x_{un},y_{un}, s_u)||p_{\theta}(z_{un}|s_u)) \Big]\\
\nonumber
&\ - \text{KL}(q_{\phi}(s_u|\{x_{un}, y_{un}\}_{n=1}^{N_u})||p(s_u)).
\]
Figure~\ref{fig:cvae_graphical_models} (parts c and d) outlines the graphical model structures of the CGM-SVAE and its variational distributions $q_\phi$.

To adapt the recognition network $r((x, y), \phi)$ used in the local inference objective (eq. \ref{eq:svaesurr}) to our model structure, we write
\vspace{-5mm}
\begin{multline}
q(s_u|\{x_{un}, y_{un}\}_{n=1}^{N_u}; \phi) \propto \\
\nonumber \textstyle
p(s_u) \exp\bigg\{\big\langle\log \pi +  \sum_{n=1}^{N_u} \log r(y_{un}, x_{un}; \phi), \; t_s(s_u) \big\rangle \bigg\},
\end{multline}
where $t_s(s_u)$ denotes the one-hot vector encoding of the mixture component index $s_u$.
That is, for each user image $x_{un}$ and corresponding set of slider values $y_{un}$, the recognition network produces a potential over the user's latent mixture component $s_u$.
These image-by-image guesses are then combined with each other and with the prior to produce the inferred variational factor on $s_u$.

This recognition network architecture is both natural and convenient. It is natural because a powerful enough $r$ can set
$r_k(y_{un}, x_{un}; \phi) \propto p_\theta(y_{un} | x_{un}, s_u=k)$,
in which case $q_\phi(s_u | \{x_{un}, y_{un}\}^{N_u}_{n=1}) \equiv p(s_u | \{x_{un}, y_{un}\}^{N_u}_{n=1})$ and there is no approximation error. It is convenient because it analyzes image-edit pair independently, and these evidence potentials are combined in a symmetric, exchangeable way that extends to any number of user images $N_u$.

\section{Experiments}
We evaluate our models and several strong baselines on three datasets. We focus on the photo editing software Adobe Lightroom. The datasets that we use cover three different types of users that can be roughly described as 1) casual users who do not use the application regularly, 2) frequent users who have more familiarity with the application and use it more frequently 3) experts who have more experience in editing photos than the other two groups. We split all three datasets into 10\% test, 10\% validation, and 80\% train set. 

\paragraph{Datasets} The casual users dataset consists of 345000 images along with the slider values that a user has applied to the image in Lightroom. There are 3200 users in this dataset. Due to privacy concerns, we only have access to the extracted features from a convolutional neural network (CNN) applied to the images. Hence, each image in the dataset is represented by a 1024-dimensional vector. For the possible edits to the image, we only focus on 11 basic sliders in Lightroom. Many common editing tasks boil down to adjusting these sliders. The 11 basic sliders have different ranges of values, so we standardize them to all have a range between $-1$ and 1 when training the model.

The frequent users dataset contains 45000 images (in the form of CNN features) and their corresponding slider values. There are 230 users in this dataset. These users generally apply more changes to their photos compared to the users in the casual group. 

Finally, the expert users dataset  (Adobe-MIT5k, collected by \citet{bychkovsky2011learning}) contains 5000 images and edits applied to these images by 5 different experts, for a total of 25000 edits.

\begin{figure}[t!]
\begin{center}
\begin{tabular}{c}
\hspace{-5mm}{\includegraphics[trim = -1mm 0mm 0mm 0mm, clip, scale = 0.6]{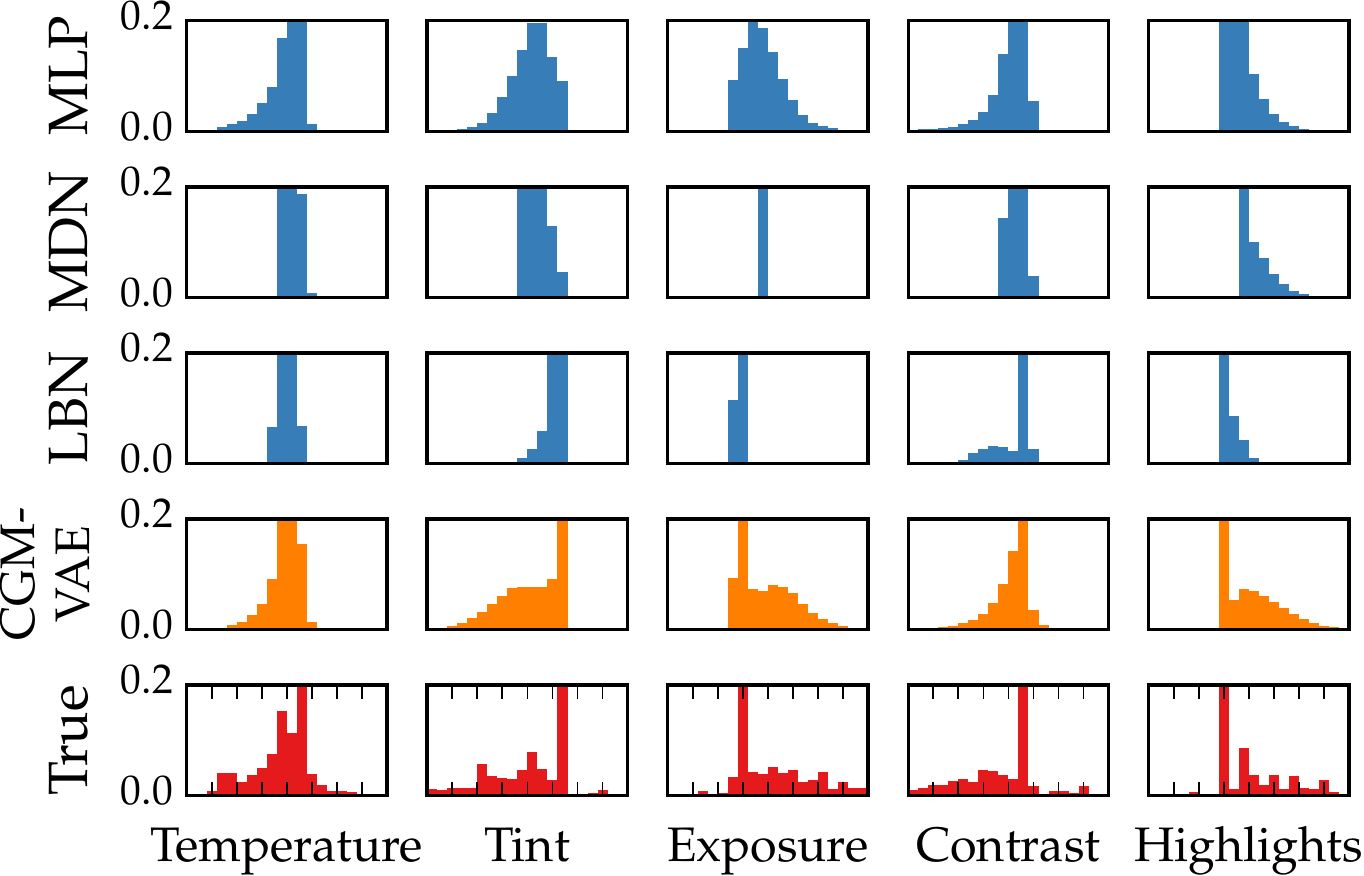}} 
\end{tabular}
\end{center}
\vspace{-4mm}
\caption{\footnotesize{Marginal statistics for the prediction of the sliders in the casual users dataset (test set). Due to space limitation, We only display the top 5 mostly used sliders in the dataset. LBN has limited success compared to CGM-VAE. MLP mostly concentrates around the mean edit. The quantitative comparison between different methods in terms of the distance between normalized histograms is provided in Table~\ref{tab:CVAEresults}.}}
\label{fig:marginals}
\end{figure}

We augment this dataset by creating new images after applying random edits to the original images. To generate a random edit from a slider, we add uniform noise from a range of $\pm$10\% of the total range of that slider. Given the augmented set of images, we extract the ``FC7'' features of a VGG-16 \cite{Simonyan14c} pretrained network and use the 4096-dimensional feature vector as a representation of each image in the dataset. After augmenting the dataset, we have 15000 images and 75000 edits in total. Similar to other datasets, we only focus on the basic sliders in Adobe Lightroom. 

\begin{figure*}[t!]
\begin{center}
\begin{tabular}{cc}
{\hspace{-10mm}\includegraphics[trim = 5mm 0mm 0mm 0mm, clip, scale = 0.5]{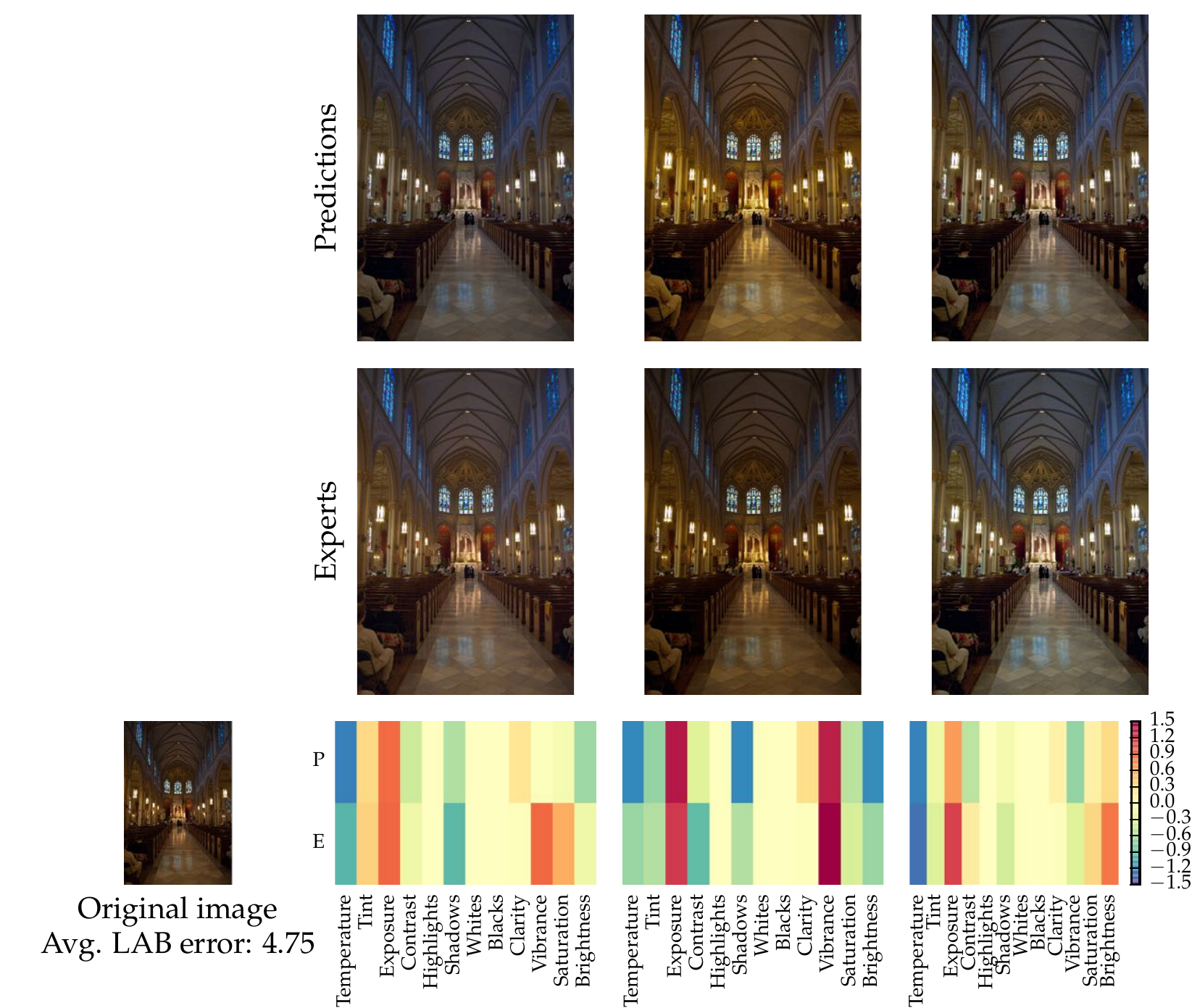}} & \hspace{5mm}
{\includegraphics[trim = 5mm 0mm 0mm 0mm, clip, scale = 0.53]{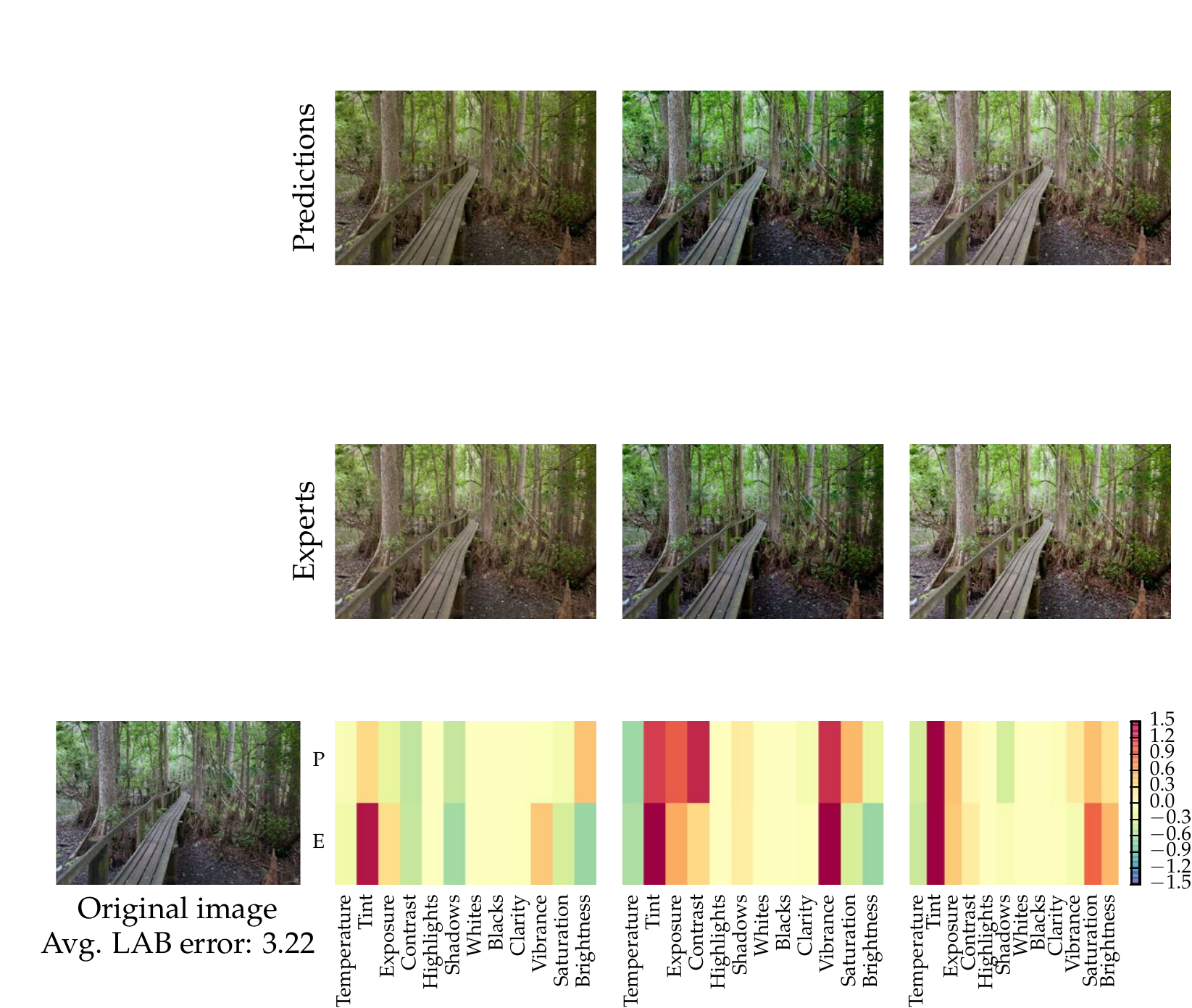}} \\
(a) & (b)
\end{tabular}
\end{center}
\vspace{-5mm}
\caption{\footnotesize{\textbf{Multimodal photo edits:} Sample slider predictions  from the CGM-VAE model (denoted by P in the figure) compared to the edits of 3 most active experts in the expert users dataset (denoted by E). The images are selected from the test subset of the dataset; the 3 samples are selected from a set of 10 proposals from the CGM-VAE model such that they align with the experts. To show the difference between the model and experts, we apply their sliders to the original image. For more examples, refer to the supplementary material.}}
\label{fig:multimodal}
\end{figure*}

\paragraph{Baselines} We compare our model for multimodal prediction with several models: a multilayer perceptron (MLP), mixture density network (MDN), and linearizing belief network (LBN). The MLP is trained to predict the mean and variance of a multivariate Gaussian distribution; this model will demonstrate the limitations of even a strong model that makes unimodal predictions. The MDN and LBN, which are specifically designed for multimodal prediction, are other baselines for predicting multimodal densities. Table \ref{tab:CVAEresults} summarizes our quantitative results. 

We use three different evaluation metrics to compare the models. The first metric is the predictive log-likelihood computed over a held-out test set of different datasets. Another metric is the Jensen-Shannon divergence (JSD) between normalized histograms of marginal statistics of the true sliders and the predicted sliders. 
Figure~\ref{fig:marginals} shows some histograms of these marginal statistics for the casual users.  

Finally, we use the mean squared error in the CIE-LAB color space between the expert-retouched image and the model-proposed image. We use the CIE-LAB color space as it is more perpetually linear compared to RGB. We only calculate this error for the experts dataset (test set) since that is the only dataset with available retouched images. To compute this metric, we first apply the predicted sliders from the models to the original image and then convert the generated RGB image to a LAB image. For reference the difference between white and black in CIE-Lab is 100 and photos with no adjustments result in an error of 10.2 . Table \ref{tab:CVAEresults}, shows that our model outperforms the baselines across all these metrics. 

\paragraph{Hyperparameters} For the CGM-VAE model, we choose the dimension of the latent variable from \{2, 20\} and the number of mixture components from the set \{3, 5, 10\}. For the remaining hyperparameters see the supplementary materials.

\begin{figure*}[t!]
\begin{center}
\begin{tabular}{c}
{\includegraphics[trim = 0mm 0mm 0mm 0mm, clip, scale = 0.47]{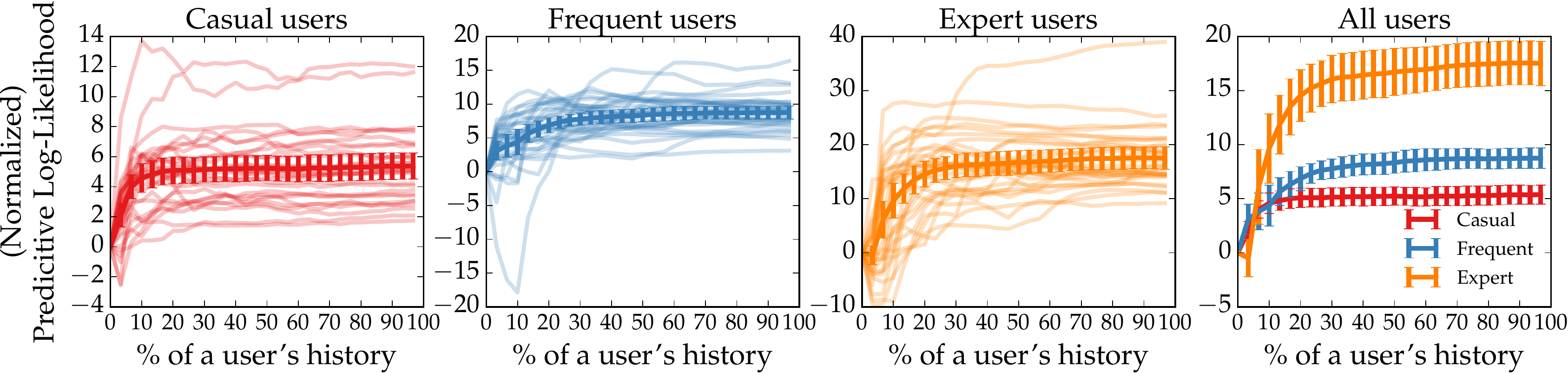}} 
\end{tabular}
\end{center}
\vspace{-4mm}
\caption{\footnotesize{Predictive log-likelihood for users in the test set of different datasets. For each user in the test set, we compute the predictive log-likelihood of 20 images, given 0 to 30 images and their corresponding sliders from the same user. 30 sample trajectories and the overall average $\pm$ s.e. is shown for casual, frequent and expert users. The figure shows that knowing more about the user (up to around 10 images) can increase the predictive log-likelihood. The log-likelihood is normalized by subtracting off the predictive log-likelihood computed given zero images. Note the different y-axis in the plots. The rightmost plot is provided for comparing the average predictive log-likelihood across datasets.}}
\label{fig:pred_LL}
\end{figure*}

\paragraph{Tasks} In addition to computing the predictive log-likelihood and JSD over the held-out test sets for all three datasets, we consider the following two tasks: 
\begin{enumerate}
\item Multimodal prediction: We predict different edits applied to the same image by the users in the experts dataset. Our goal is to show that CGM-VAE is able to capture different styles from the experts.
 
\item Categorizing the users and adapting the predictions based on users' categories: We show that the CGM-SVAE model, by clustering the users, makes better predictions for each user. We also illustrate how inferred user clusters differ in terms of edits they apply to similar images.

\end{enumerate}

\subsection{Multimodal predictions} 
To show that the model is capable of multimodal predictions, we  
propose different edits for a given image in the test subset of the experts dataset. To generate these edits, we sample from different cluster components of our CGM-VAE model trained on the experts dataset. For each image we generate 20 different samples and align these samples to the experts' sliders. From the 5 experts in the dataset, 3 propose a more diverse set of edits compared to the others; hence, we only align our results to those three to show that the model can reasonably capture a diverse set of styles.  

For each image in the test set, we compare the predictions of MLP, LBN, MDN and the CGM-VAE with the edits from the 3 experts. In MLP (and also MDN), we draw 20 samples from the Gaussian (mixture) distribution with parameters generated from the MLP (MDN). For the LBN, since the network has stochastic units, we directly sample 20 times from the network. We align these samples to the experts' edits and find the LAB error between the expert-retouched image and the model-proposed image.

To report the results we average across the 3 experts and across all the test images. The LAB error in Table \ref{tab:CVAEresults} indicates that CGM-VAE model outperforms other baselines in terms of predicting expert edits. Some sample edit proposals and their corresponding LAB errors are provided in Figure~\ref{fig:multimodal}. This figure shows that the CGM-VAE model can propose a diverse set of edits that is reasonably close to those of experts. For further examples see the supplementary material.

\begin{figure*}
\begin{center}
\begin{tabular}{cc}
{\hspace{-10mm} \includegraphics[trim = 5mm 0mm 0mm 0mm, clip, scale = 0.33]{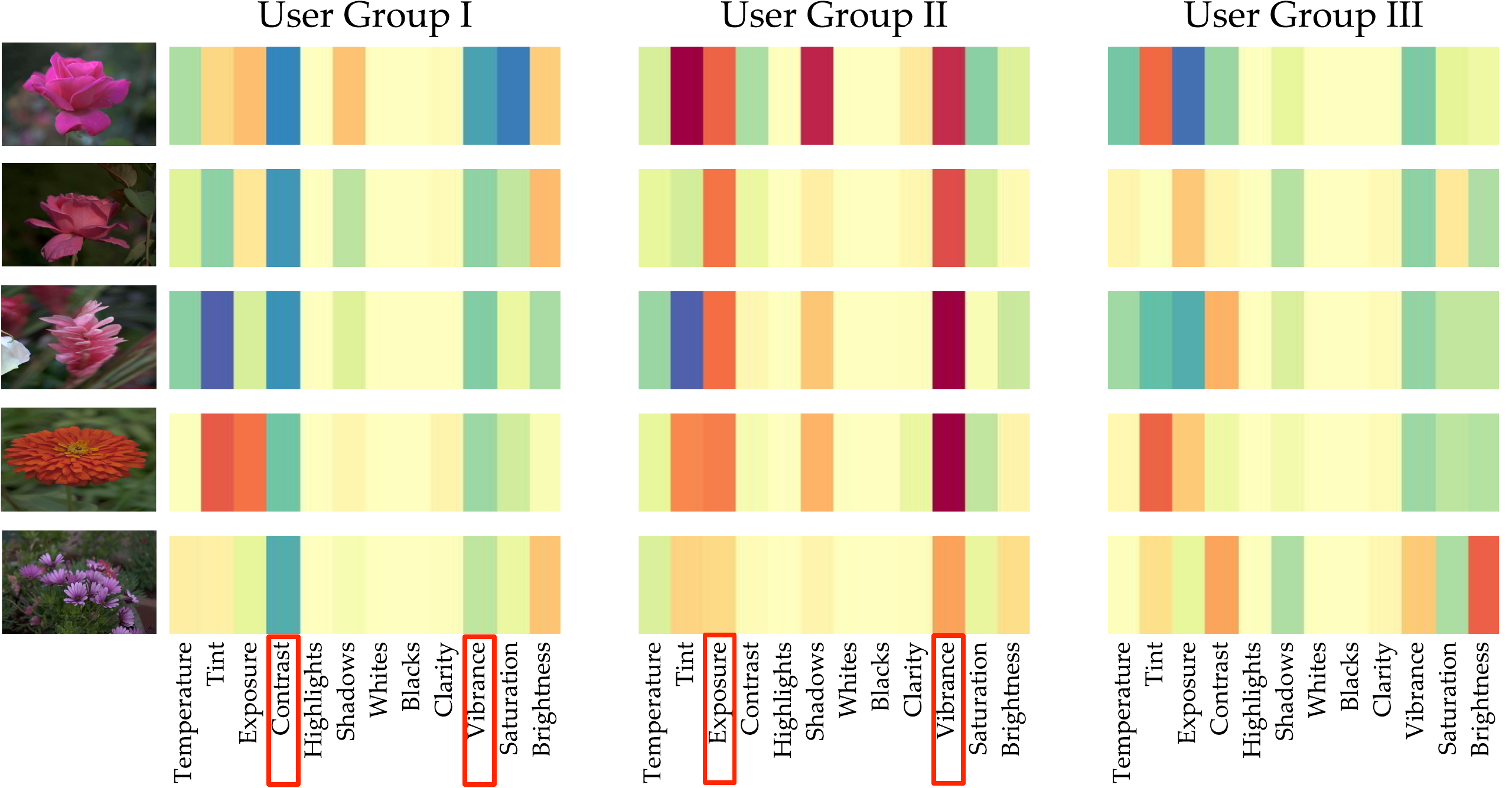}} & \hspace{5mm}
{\includegraphics[trim = 5mm 0mm 0mm 0mm, clip, scale = 0.33]{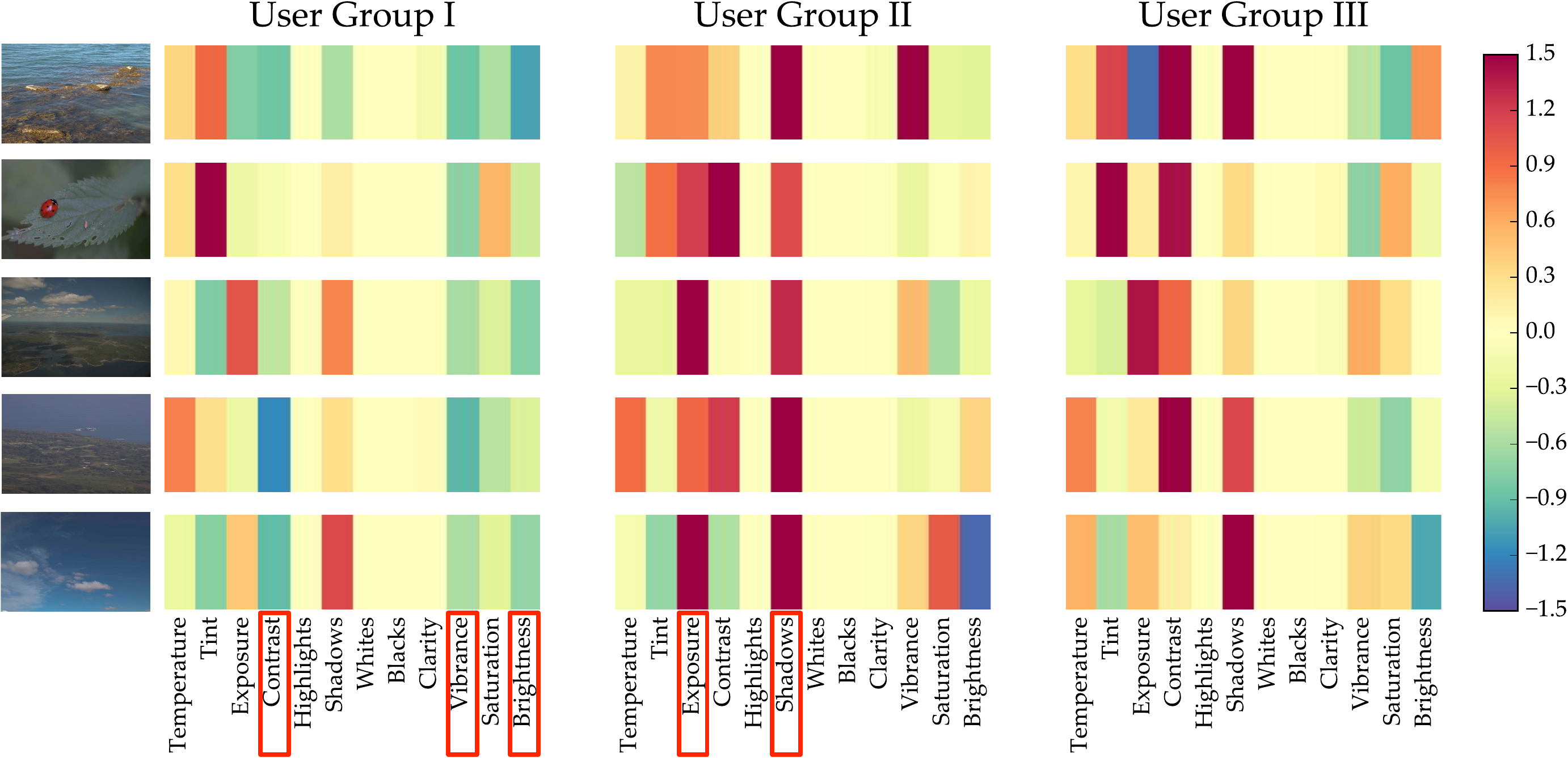}} \\
(a) & (b)
\end{tabular}
\end{center}
\vspace{-5mm}
\caption{\footnotesize{\textbf{User categorization:} Two examples of sample edits for three different user groups which the CGM-SVAE model has identified (in the experts dataset). (a) For similar flower photos, users in group I prefer to use low contrast and vibrance, whereas group II users tend to increase the exposure and vibrance from their default values. There is also group III users which do not show any specific preference for similar flower photos. (b) The same user groups for another set of similar photos with dominant blue colors. For more examples, see the supplementary materials.}}
\label{fig:categorization}
\end{figure*}

\begin{table*}[h!]
 \begin{adjustwidth}{-5mm}{}
\begin{center}
\begin{tabular}{L{50pt}C{50pt}C{50pt}C{50pt}C{50pt}C{50pt}C{50pt}C{50pt}C{50pt}@{}} 
\hline
\hspace{1pt}
\textbf{Dataset} & \multicolumn{2}{c}{\textbf{Casual}} & \multicolumn{2}{c}{\textbf{Frequent}} & \multicolumn{3}{c}{\textbf{Expert}} \\
\hline
\hspace{2 pt}\textbf{Eval. Metric}  & {LL} & {JSD} & {LL} & {JSD} & {LL} & {JSD} & {LAB} \\
\hline
\vspace{1mm}
MLP & $-15.71 \pm 0.21$ & $0.26 \pm 0.04$ &  $-2.72 \pm 0.31$ &  $0.11 \pm 0.02$ & $-4.28 \pm 0.12$ & $0.22 \pm 0.06$ & $7.81 \pm 0.26$\\
\vspace{1mm}
LBN &  $-7.12 \pm 0.15$ &  $0.14 \pm 0.02$ & $-3.7 \pm 0.43$ &  ${0.13 \pm 0.02}$ & ${-4.89 \pm 0.24}$& ${0.17 \pm 0.04}$ & $7.44 \pm 0.29$\\
\vspace{1mm}
MDN &  $-14.53 \pm 0.25$ &  $0.31 \pm 0.06$ & $-1.67 \pm 0.47$ &  $0.24 \pm 0.08$ & $-4.91 \pm 0.07$ & $0.28 \pm 0.11$ & $8.41 \pm 0.27$\\
\vspace{1mm}
CGM-VAE & $\mathbf{-6.39 \pm 0.11}$ & $\mathbf{0.10 \pm 0.02}$ & ${\mathbf{-1.42 \pm 0.18}}$& $\mathbf{0.08 \pm 0.02}$ &  $\mathbf{ -2.6 \pm 0.15}$ & $\mathbf{0.12 \pm 0.05}$ & $\mathbf{6.72 \pm 0.27}$\\
\hline
\end{tabular}
\end{center}
\caption{\footnotesize{\textbf{Quantitative results:} \textit{LL}: Predictive log-likelihood for our model CGM-VAE and the three baselines. The predictive log-likelihood is calculated over the test sets from all three datasets. \textit{JSD}: Jensen-Shannon divergence between normalized histograms of the true sliders and our model predictions over the test sets (lower is better). See Figure \ref{fig:marginals} for an example of these histograms. \textit{LAB}: LAB error between the images retouched by the experts and the images retouched by the model predictions. For each image we generate 3 proposals and compare that with the images generated by the top 3 active experts in the experts dataset.}} 
\vspace{-4mm}
\label{tab:CVAEresults}
\end{adjustwidth}
\end{table*}

\subsection{Categorization and personalization}
Next, we demonstrate how the CGM-SVAE model can leverage the knowledge from a user's previous edits and propose better future edits. For the users in the test sets of all three datasets, we use between 0 and 30 image-slider pairs to estimate the posterior of each user's cluster membership. We then evaluate the predictive log-likelihood for 20 other slider values conditioned on the images and the inferred cluster memberships.

Figure~\ref{fig:pred_LL} depicts how adding more image-slider combinations can generally improve the predictive log-likelihood. The log-likelihood is normalized by subtracting off the predictive log-likelihood computed given zero images. The effect of adding more images is shown for 30 different sampled users; the overall average for the test dataset is also shown in the figure. To compare how various datasets benefit from this model, the average values from the 3 datasets are overlaid. According to this figure, the frequent users benefit more than the casual users and the expert users benefit the most. \footnote{To apply the CGM-SVAE model to the experts dataset, we split the image-slider combinations from each of the 5 experts into groups of 50 image-sliders and pretend that each group belongs to a different user. This way we can have more users to train the CGM-SVAE model. However, this means the same expert may have some image-sliders in both train and test datasets. The significant advantage gained in the experts dataset might be due in part to this way of splitting the experts. Note that there are still no images shared across train and test sets.}

To illustrate how the trained CGM-SVAE model proposes edits for different user groups, we use a set of similar images in the experts dataset and show the predicted slider values for those images. Figure \ref{fig:categorization} shows how the inferred user groups edit two groups of similar images. This figure provides further evidence that the model is able to propose a diverse set of edits across different groups; moreover, it shows each user group may have a preference over which slider to use. For more examples see the supplementary material.

\section{Conclusion}
We proposed a framework for multimodal prediction of photo edits and extend the model to make personalized suggestions based on each user's previous edits. Our framework outperforms several strong baselines and demonstrates the benefit of having interpretable latent structure in VAEs. Although we only applied our framework to the data from photo editing applications, it can be applied to other domains where multimodal prediction, categorization and personalization are essential. Our proposed models could be extended further by assuming more complicated graphical model structure such as admixture models instead of the Gaussian mixture model that we used. Furthermore, the categories learned by our model can be utilized to gain insights about the types of the users in the dataset. 

\renewcommand*{\bibfont}{\small}
\bibliographystyle{abbrvnat}
\bibliography{LrVAE}

\end{document}